\documentclass[letterpaper, 10 pt, conference]{ieeeconf}  

\IEEEoverridecommandlockouts                              
\usepackage[utf8]{inputenc}
\usepackage[english]{babel}
\usepackage{xspace}
\usepackage{dirtytalk}
\usepackage{stfloats}
\usepackage{graphicx}
\usepackage{lipsum}
\usepackage{amsmath,amssymb,amsfonts}

\title{\LARGE \bf
DyOb-SLAM : Dynamic Object Tracking SLAM System
}

\author{\parbox{3 in}{\centering Rushmian Annoy Wadud\\
Master Student, School of Aerospace and Mechanical Engineering\\
University of Oklahoma\\ 
Norman, OK 73019\\
{\tt\small rushmian@gmail.com}}
\hspace*{ 0.5 in}
\parbox{3 in}{ \centering Wei Sun \\
Assistant Professor, School of Aerospace and Mechanical Engineering\\
University of Oklahoma\\ 
Norman, OK 73019\\
        {\tt\small wsun@ou.edu}}%
}

\begin{document}

\maketitle
\thispagestyle{empty}
\pagestyle{empty}

\begin{abstract}

Simultaneous Localization \& Mapping (SLAM) is the process of building a mutual relationship between localization and mapping of the subject in its surrounding environment. With the help of different sensors, various types of SLAM systems have developed to deal with the problem of building the relationship between localization and mapping. A limitation in the SLAM process is the lack of consideration of dynamic objects in the mapping of the environment. 
We propose the Dynamic Object Tracking SLAM (DyOb-SLAM), which is a Visual SLAM system that can localize and map the surrounding dynamic objects in the environment as well as track the dynamic objects in each frame. With the help of a neural network and a dense optical flow algorithm, dynamic objects and static objects in an environment can be differentiated. DyOb-SLAM creates two separate maps for both static and dynamic contents. For the static features, a sparse map is obtained. For the dynamic contents, a trajectory global map is created as output. As a result, a frame to frame real-time based dynamic object tracking system is obtained. With the pose calculation of the dynamic objects and camera, DyOb-SLAM can estimate the speed of the dynamic objects with time. 
The performance of DyOb-SLAM is observed by comparing it with a similar Visual SLAM system, VDO-SLAM and the performance is measured by calculating the camera and object pose errors as well as the object speed error. 

\end{abstract}

\section{INTRODUCTION}

The Simultaneous Localization and Mapping \cite{SLAM} problem can be considered as maintaining a mutual relationship between the mapping and localization of a robot in an unexplored environment. Without mapping, the subject cannot be localized and without the pose estimation of the subject, the map cannot be formed. With the help of the sensors, the significant landmarks or key-features can be located, which will be processed by the device to match and link them with the previously observed landmarks, as well as store them for mapping purpose. The state and position of a robot can be estimated after updating the landmark features and can be used for mapping as well. Many sensors have been utilized in SLAM, such as laser range sensors, rotary encoders, inertial sensors, GPS, and cameras. Depending on the sensors, SLAM can be classified into various types. In this paper, we will focus on Visual SLAM based on camera sensors. 
\par 
\begin{figure}[]
    \centering
    \includegraphics[scale=0.092]{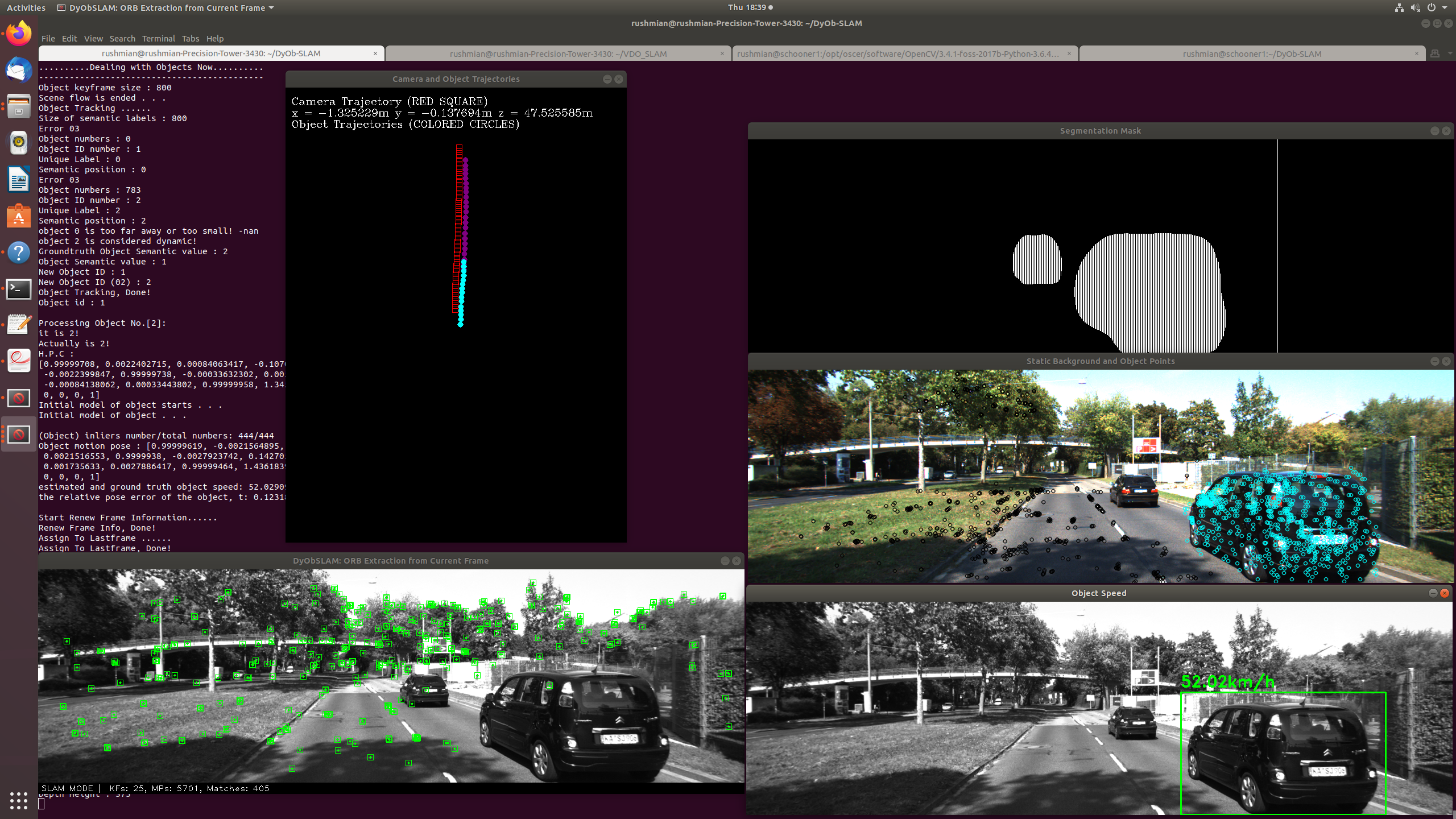}
    \caption{Output of DyOb-SLAM}
    \label{figure output-dyob}
\end{figure}

The main objective of Visual SLAM is to estimate the camera trajectory and reconstruct the surrounding environment as a map. Most of Visual SLAM algorithms are based on an assumption called \say{Scene Rigidity Assumption} or the static world assumption. This assumption has been developed in many approaches of SLAM systems where it is considered that the environment does not contain dynamic objects and is completely static. While the scene rigidity assumption is required for the ease of computation since dealing with dynamic objects is computationally expensive, it creates a limitation for the real world based applications of Visual SLAM. SLAM systems developed in recent years start to take dynamic objects into consideration. These algorithms function mostly in two ways: 
\begin{enumerate}
    \item The moving objects detected from the sensors are treated as outliers and removed from the estimation process.
    \item The moving objects are tracked separately using multi-target tracking approaches after they are detected.
\end{enumerate}

DynaSLAM \cite{dynaslam} functions in the former way, where it detects \say{prior} dynamic objects, i.e. objects which are potentially dynamic, and then segment the objects using a Mask-RCNN model before removing the segmented portions from the frames. The map generated by DynaSLAM is based on the static objects in the surrounding environment. VDO-SLAM \cite{vdo-slam}, on the other hand, is a system which functions in the latter way, i.e. it tracks the dynamic objects and estimates the object poses (both static and dynamic). 
\par

The Dynamic Object Tracking SLAM (DyOb-SLAM) we propose in this paper is a combination of DynaSLAM and VDO-SLAM. Figure \ref{figure output-dyob} shows the output of the DyOb-SLAM system. Different features are mentioned below:   
\begin{itemize}
    \item Our system consists of a Mask-RCNN module for segmenting out the dynamic objects based on prior, and for a better segmentation result, the Multi-View geometric segmentation algorithm from DynaSLAM has been added.
    \item For getting a robust system of dynamic object tracking, the optical flow and scene flow algorithms are implemented.
    \item The back-end of our system consists of Bundle Adjustment feature for the static object points, the Partial Batch Optimization module for creating local maps and a Full Batch Optimization module for the final result - a global map.
    \item With the help of the Bundle Adjustment algorithm, a sparse map can be obtained with the static feature points. 
\end{itemize}

The following outputs can be obtained: 
    \begin{enumerate}
        \item A current frame showing the ORB features along with mask information and object labels.
        \item A sparse map based on the static features.
        \item A global map showing the dynamic contents and their motion updated with time.
    \end{enumerate}  
The rest of the paper is organized as follows. The literature review is discussed in Section \ref{sec:lit_review}. The methodology of DyOb-SLAM is discussed in Section \ref{sec:methodology}. The experimental data and results are presented in Section \ref{sec:exp}. Section~\ref{sec:Conclusion} provides a summary of the results.

\section{Literature Review} \label{sec:lit_review}
In SLAM methods, the most common assumption made is that the observed scene in the environment is static occluding out the dynamic contents from calculation. In earlier designs of SLAM system \cite{ptam} \cite{orb-slam} \cite{orb-slam2} \cite{kinectfusion}, this assumption was established. In \cite{orb-slam} and \cite{orb-slam2}, dynamic object points are considered as outliers and sparse maps of only static features are obtained. In a system where there is no moving object, any motion of an object is treated as an outlier and thus occluded from the tracking and mapping \cite{detect-fusion}. This will result in failure of tracking and subsequently the mapping in many realistic scenarios. For this purpose, dynamic object detection is widely utilized currently to deal with this issue.    

\par 
Some of the SLAM systems use object detection algorithm to form a semantic map of the environment \cite{slam++} \cite{cnn-slam} \cite{semantic-fusion}, but the dynamic object features are not taken into calculation and occluded in the tracking process. Similar to these systems, in Detect-SLAM \cite{detect-slam}, ORB-SLAM2 \cite{orb-slam2} and the Single Shot Detector (SSD) \cite{ssd} co-exist in the system to detect the dynamic objects for occlusion in the local map, and with the help of SSD, an instance level semantic map is formed based on the static objects. ORB-SLAM2 creates a sparse reconstruction of the environment from the extracted ORB (Oriented FAST and Rotational BRIEF) \cite{orb} features and SSD produces discretized bounding boxes and generates scores for presence of each object category in each default box. With the help of the single-layered convolutional layer YOLOv3 \cite{yolov3}, the dynamic objects which gets occluded in the tracking stage, are detected in \cite{detect-fusion} creating an instance level segmentation using 3D geometric segmentation method. In \cite{mask-fusion}, a purely geometric map of the static scene is constructed by occluding the features detected from the motion disturbances in the static scene. 

\par  
DynaSLAM \cite{dynaslam} is a system which is built on ORB-SLAM \cite{orb-slam} and ORB-SLAM2 \cite{orb-slam2} and can be implemented with monocular, stereo and RGB-D image sets. The system is comprised of a neural network Mask-RCNN \cite{mask-rcnn} to segment out the dynamic objects along with a multi-view Geometry algorithm, a tracking component to track the static objects and a mapping component to map out the static feature points. This system has a dynamic object occlusion algorithm, along with a background inpainting algorithm to fill in the occluded spaces in the frames with previously observed background. On the other hand, systems like DOT or Dynamic Object Tracking \cite{dot} combine instance segmentation and multi-view geometry to generate masks for dynamic objects, which works like a tracker. Visual Dynamic Object-aware SLAM (VDO-SLAM) system  \cite{vdo-slam} is the first dynamic SLAM system to perform motion segmentation. The system's novelty is that it can track multiple dynamic objects with the semantic information, estimate the camera pose of both static and dynamic structures and it has an object velocity extracting algorithm. The tracking of multiple dynamic objects is done using a dense optical flow algorithm which propagates a unique object ID assigned to the points in the segmented regions.
\par 
Although DynaSLAM is a better performing SLAM system, occlusion of dynamic features and background inpainting characteristics are not very reliable for real world based applications. On the other hand, VDO-SLAM fails to produce real time based outputs due to the pre-processing of the frames. Both of the systems use deep learning in SLAM systems to produce semantic information which needs a lot of energy for computation and cannot be run efficiently in real time. One solution to improve a dynamic object based SLAM system by producing better computational speed and energy is cloud computation. The higher computations are done in the cloud to get real time data for building up the map. In robotics, there are many cloud computation platforms for robots like - Rapyuta \cite{rapyuta} which is an open source Platform-as-a-Service (PaaS) framework designed specifically for robotics applications (SLAM), DAvinCi \cite{davinci} which is a software framework that provides the scalability and parallelism advantages of cloud computing for service robots in large environments, ROS-bridge which bridges communication between a robot and a single ROS (Robot Operating System) \cite{ros} environment in the cloud. 

\section{Methodology} \label{sec:methodology}
\begin{figure*}[t]
    \centering
    \includegraphics[scale=0.3]{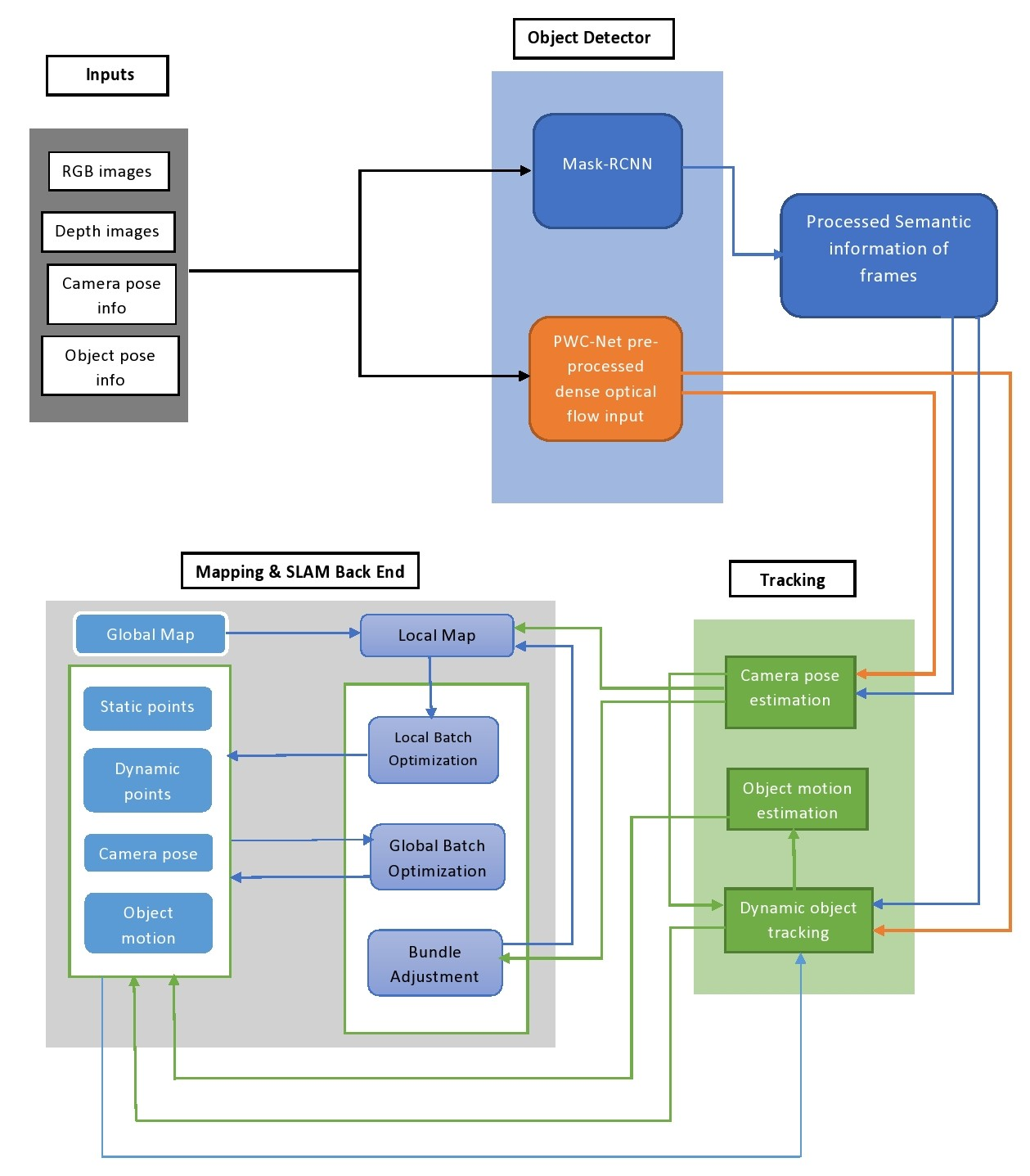}
    \caption{Block diagram of DyOb-SLAM}
    \label{figure 02}
\end{figure*}

DyOb-SLAM is a system which tracks dynamic objects, maps both the static and dynamic objects separately and simultaneously estimates the camera and object poses by comparing with the ground truth information. The system is comprised of:  
\begin{itemize}
    \item an object detector module
    \item a tracking component
    \item two different mapping algorithms
    \item an orientation optimizing back-end
\end{itemize}
The input to the system are stereo, RGB-D images. Figure \ref{figure 02} shows the block diagram of the DyOb-SLAM system.
\par
The object detector module at first creates instance-level semantic segmentation information of the dynamic objects present in each frames. The dynamic object segmentation is based on \emph{prior}, i.e. the objects which are potential dynamic objects or movable in real world, for example - car, people, etc. Due to the semantic segmentation of \emph{a priori} dynamic contents, the static and dynamic objects are separated which will be easier to track the objects separately. With the help of a dense optical flow algorithm, the number of dynamic objects to be tracked is maximized. The dense optical flow information is pre-processed using PWC-Net, which samples all the points of dynamic contents in the frame from the semantic information. 
\par 
The semantic information along with the optical flow information are then passed onto the tracking module which tracks the dynamic points extracted from the semantic and optical flow information and produces camera and object pose information. It also compares the pose information with the ground truth information provided and calculates the pose errors which are obtained as output. Next, the tracking information is projected into two different maps - a sparse point-cloud map for the static contents and a global map for the dynamic contents along with the camera which provides trajectory information along each frames. The sparse map is optimized using a Bundle Adjustment algorithm and the global map is optimized using batch optimization process (both full and partial). The different stages are described in the subsections below.

\subsection{Object Detector}
\subsubsection{Mask-RCNN}
In the system, a Convolutional Neural Network (CNN) is used to segment out the potential dynamic objects from the frame. An instance-level semantic segmentation algorithm module, Mask-RCNN is used which is an extended version of Faster-RCNN with an added branch for predicting an object mask in parallel with the bounding box feature. It can extract both pixel-wise semantic segmentation and instance labels of objects. In this system, both the functions are used - the \emph{priori} dynamic objects are segmented out and instance labels are obtained to track the dynamic pixels. 
\par
The input to the Mask-RCNN module are stereo, RGB-D images. The network has been trained in such a way that it can detect different potential dynamic objects, such as people, car, trains, truck, birds, dog, cat, etc. The network are trained on MS COCO to segment out the selected classes. The main concept is to segment out these classes and obtain an output matrix of size $m\times n \times l$ where m,n are the row and column of the input matrix of images and l is the number of objects in the frames. For each class of objects, a specific ID value has been assigned which is obtained in the output matrix where the pixel has been masked by the neural network. For each value in the matrix, a specific color of mask is assigned for visualizing dynamic object segmentation in the tracking output (see figure \ref{figure 03}).
\par
\begin{figure}[]
    \centering
    \includegraphics[scale=0.2]{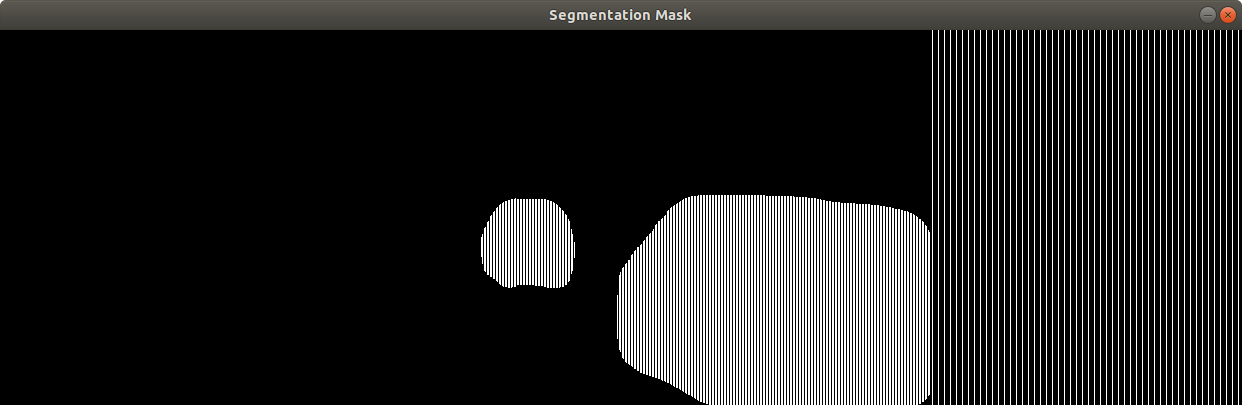}
    \caption{Semantic Segmentation of scene using Mask-RCNN in the experiment}
    \label{figure 08}
\end{figure}

  
The figure \ref{figure 08} shows the semantic segmentation of a frame scene. The dataset that is used to test the system contains cars and other vehicles, which are the only objects segmented out. The ground truth of the dataset contains the object pose and semantic information. The Mask-RCNN's output matrix is then processed for the other modules (Tracking and Mapping) in such a way that the dynamic objects detected are relabelled according to the ground truth. As a result the output becomes closer to ground truth information.   

\par 
\subsubsection{Dense Optical Flow}
PWC-Net, a dense optical flow algorithm is used to at first pre-process the optical flow information of the input images. These pre-processed optical information are used as input to the system. The dense optical flow information sample all the points from the dynamic objects within the segmented masks. This helps to maximize the number of tracked points and later on used for tracking multiple objects. Even if the semantic segmentation fails at one point, dense optical flow information can help obtain the object masks again by tracking the unique ID for each points in the mask around the object. Since sparse feature matching method is not very effective for tracking dynamic contents in the long term consecutive frames, optical flow estimation has been used for that purpose. 

\subsection{Tracking}
The input to the tracking component are the RGB images, the depth information of each frames, the segmentation masks and the optical flow information obtained from the object detector module. Multiple functions take place simultaneously in the tracking component and they are divided in 3 modules : 
\begin{itemize}
    \item ORB Feature Extraction
    \item Camera Pose Estimation
    \item Object Motion Tracking 
\end{itemize}

\begin{figure}[]
    \centering
    \includegraphics[scale=0.3]{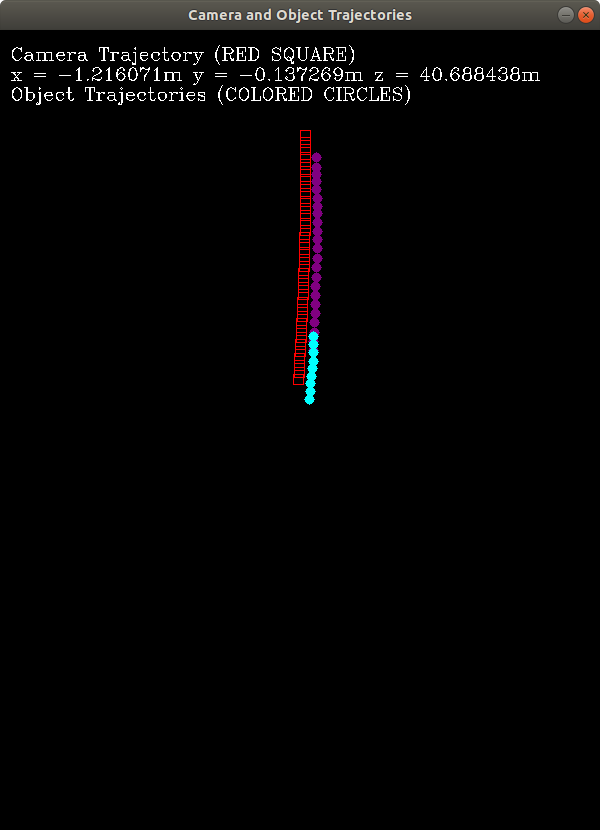}
    \caption{Tracked Dynamic Objects}
    \label{figure 03}
\end{figure}

\subsubsection{ORB Feature Extraction}
This module is similar to the process used in ORB-SLAM or ORB-SLAM2. This module consists of the following aspects,
\begin{itemize}
    \item \textbf{Localization:} It localizes the camera, finding feature matches in every frame and forms visual odometry tracks of unmapped regions.
    \item \textbf{Loop Closing:} It uses a place recognition algorithm to detect and validate large loops. For place recognition, a Bag of Words module DBoW2 \cite{DBoW2} is used.
\end{itemize}

\begin{figure}[]
    \centering
    \includegraphics[scale=0.19]{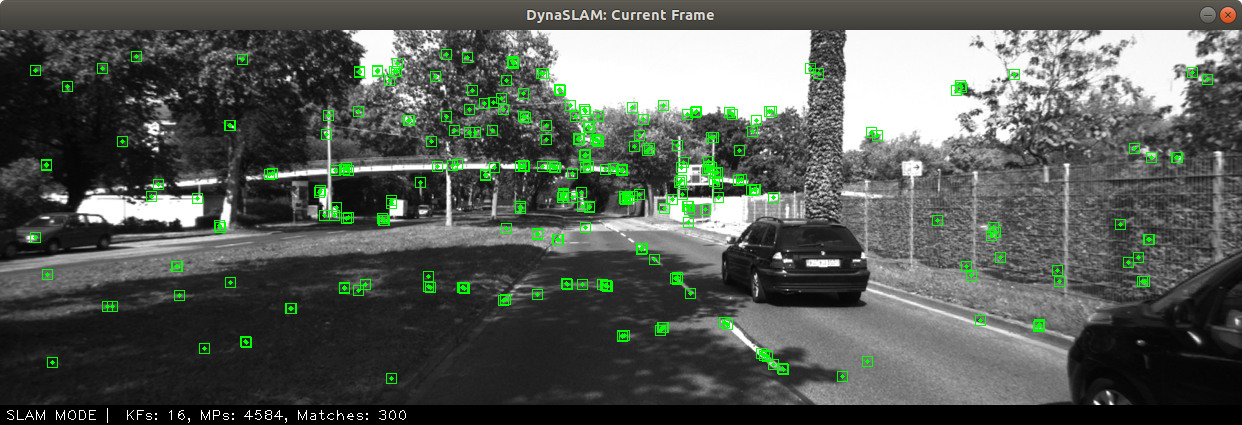}
    \caption{ORB feature extraction in the current frame}
    \label{fig orb}
\end{figure}

Figure \ref{fig orb} shows the output of ORB feature extraction. It detects the \emph{corner} features in the frames, i.e. ORB features and extract them to form a sparse map as output. The ORB features are extracted from the static part of the image frames excluding the segmented mask portions of dynamic objects.

The ORB features are used for tracking, place recognition (Loop Closing) and local mapping functions and these features are very robust to rotation and scale \cite{orb} \cite{orb-slam2}. ORB, which is in short for Oriented FAST and Rotated BRIEF, is a combination of FAST (Features from Accelerated Segment Test) keypoint detection method and BRIEF (Binary Robust Independent Elementary Features) descriptor which uses binary test for smoothing the noisy patches of pixels.
\par 
Since the system deals with stereo and RGB-D image types, the ORB extraction is done for both the images (left and right) and the ORB extractor searches for an ORB match in both the images. This matched keypoint associated with the depth information are used for differentiating between close and far points. Keypoints are considered close points according to [5] when the associated depth is less than 40 times than the stereo/RGB-D baseline. Otherwise, the keypoint will be considered a far point. The baseline has been calculated according to \cite{Stereo-RGBD-baseline}. The close points are usually considered and triangulated while the far points are discarded. But the far points are triangulated if they are viewed from multiple view points.

\subsubsection{Camera Pose Estimation}
After the sparse and dynamic features are separated using the object detector module, the camera pose is estimated using the static feature points. For initializing the process, motion models are generated to compare the inlier numbers depending on the camera reprojection error. Two models are formed for robust estimation - one is used by considering the previous camera motion and the other produces a new motion transform using PnP based RANSAC algorithm. Each of the models creates a number of inliers and the model with the most inliers is chosen for initialization.
\par 
For camera pose estimation, the reprojection error equation is at first established. If $P_{k-1}$ is a set of static points at frame (k-1) in the global reference frame and $C_{k}$ is the set of corresponding static feature points in the image at frame k, then the camera pose $X_k$ according to \cite{vdo-slam} is estimated by minimizing the reprojection error - 
\begin{equation} \label{eq:1}
    e(X_k) = C_k - \pi ({X_k}^{-1} P_{k-1})
\end{equation}
Here $\pi (.)$ is a projection function. A least squares error function is established from equation \ref{eq:1} using Lie-Algebra parameterization of SE(3). This least squares error function is then minimized using the Levenberg-Marquardt algorithm \cite{Levenberg-Marquardt}.  

\subsubsection{Object Motion Tracking}
The Mask-RCNN and optical flow modules segment out the potentially dynamic objects separating the static and dynamic features. For updating the dynamic object segmentation information, an optical flow algorithm called scene flow algorithm is executed. Using the scene flow algorithm, the motion of the dynamic objects are calculated. This algorithm helps to further detect the dynamic objects properly, i.e. can decide whether an object is in motion. Since the scene flow estimation of static objects is zero, a threshold is selected to decide whether the object is static or dynamic. If the magnitude of a scene flow vector of a certain point in the frame is greater than the threshold, that point is considered as dynamic. The scene flow vector is calculated using the camera pose and the motion of the object's point between two consecutive frames. 
\par 
With the help of the dense optical flow information, a point label which is a unique object identifier is associated with the dynamic object points. If the first dynamic object is detected, the point label will read $l=1$ where $l\in \mathcal{L}$ and $\mathcal{L}$ is a fixed tracking label set. For static objects and background, the value of $l$ is considered to be 0. So for frame k, the point labels will be aligned with the corresponding point labels obtained in previous frame k-1. 
\par
Similar to the camera pose estimation, object pose estimation is also calculated at first by finding the reprojection error and then Lie-Algebra parameterization of SE(3). If the object point motion from frame k-1 to k in the global reference frame is $^{k-1}{O_{k}}$, the motion estimation equation can be derived as : 
\begin{equation} \label{eq:2}
    P_k =  {^{k-1}{O_{k}}} P_{k-1}
\end{equation}
This equation is the point motion estimation equation. Here, $P_k$ is the set of static points in the frame k of image and $P_{k-1}$ is the static points in the frame $k-1$. Using equation \ref{eq:2} the reprojection error between the object point in global reference frame and the static points in image frame : 
\begin{equation} \label{eq:3}
    e(^{k-1}{O_{k}}) = C_k - \pi ({X_k}^{-1} [^{k-1}{O_{k}}] P_{k-1})
\end{equation}
After Lie Algebra parameterization of SE(3) the optimal solution is obtained minimizing the least squares error function.
\par 
With the help of the dynamic object motion tracking, the object speed is also calculated from the difference of the estimated speed $v_e$ and ground truth speed $v_g$, i.e., 
\begin{equation} \label{eq:4}
    E = v_g - v_e
\end{equation}

\subsection{Mapping}
The mapping component produces two types of map - Sparse Map for the static features and Global Map for the dynamic contents and camera motion.

\subsubsection{Sparse Map}
\begin{figure}[]
    \centering
    \includegraphics[scale=0.2]{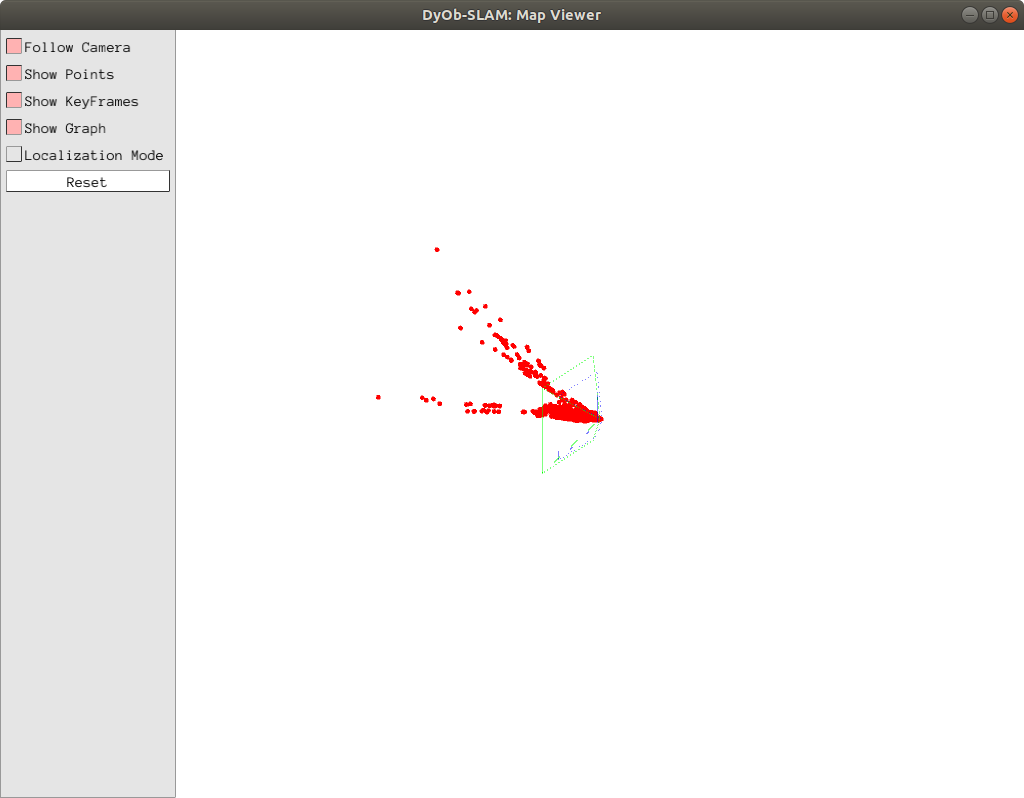}
    \caption{Sparse map of the static features in the scene}
    \label{figure 04}
\end{figure}

The ORB features \cite{orb-slam} \cite{orb-slam2} extracted in the tracking component are used to produce a sparse point-cloud map of the static background. This sparse map is also a local map consisting of triangulated ORB features from connected keyframes. The corresponding ORB features are matched with the previous keyframes and new points are generated into the local map. For triangulation of the ORB matches, the parallax error, reprojection error and scale consistency are checked. Figure \ref{figure 04} shows the sparse map obtained from the ORB extraction.   

\subsubsection{Global Map}
The inputs to the formation of global map are the output from the tracking component, i.e. the camera pose information and the object motion. With each frame and gradual change in time step, the detected object motion and camera pose are saved and continuously updated. The inlier points obtained from the previous frames are utilized to gather the track correspondences in the current frame to estimate the camera pose and object motion. With different assigned colors for different values in the pixel matrices, the camera and object trajectories are visualized in the global map. It is only based on the dynamic content and camera. Figure \ref{figure 03} shows the global trajectory map of the detected dynamic objects.

\subsection{SLAM Back-end}
The SLAM back-end is the part of the system where the data and output obtained from the other modules are optimized to get a better and more optimized output as a whole. DyOb-SLAM uses the $g^{2}o$ \cite{g2o} module for all the optimization functions and the Levenberg-Marquardt \cite{Levenberg-Marquardt} method is implemented from it to locate a local minimum of the multivariate function for the pose error that is expressed as the sum of squares of non-linear, real-valued functions. The SLAM system is designed using three different types of optimization techniques for optimized static featured map, camera pose error and dynamic object motion estimation - 
\begin{itemize}
    \item Bundle Adjustment
    \item Local Batch Optimization
    \item Global Batch Optimization
\end{itemize}

\subsubsection{Bundle Adjustment (BA)}
The selected keyframes and sparse map points obtained from the mapping component are optimized by the Bundle Adjustment. It is used only for optimization of different attributes of the static features. The Levenberg-Marquadt method is used for Bundle Adjustment which is implemented in g2o. 
\par 
Three different types of Bundle Adjustment are used in the system. To optimize the camera pose estimation obtained from the tracking component, a {motion-only bundle adjustment} is performed. This optimizes the camera orientation and minimizes the reprojection error obtained from the matched keypoints. {Local bundle adjustment} is used to optimize the selected local keyframes and local map points of the static features. After loop closure, {Full bundle adjustment} is utilized to optimize all the map points and  keyframes except the origin keyframe to achieve the optimal solution.

\subsubsection{Local Batch Optimization}
Local Batch Optimization is used for formation of the local map and to be used as input of Global Batch Optimization. Its purpose is to ensure correct camera pose which is assisted by the Bundle Adjustment algorithm and sent to the global batch optimizer to form a precise global map. It optimizes the camera pose estimation by minimizing the reprojection error using the Levenberg-Mardquardt method. For local optimization, only the static features are optimized as dynamic contents are big constraints.   

\subsubsection{Global Batch Optimization}
After local batch optimization, its output along with the output of tracking components are directly used to optimize the global map. The tracked object points are optimized fully to form the global map with every consecutive time steps and obtain updated object poses. This optimization minimizes the pose error of both camera and objects. The global map is obtained after all the time steps and frames are processed and the pose estimations are globally optimized.

\subsection{Cloud Computing}
Cloud computing helps such SLAM systems with object detection algorithms to operate in real time, as these algorithms require high computational power, such as high end GPUs (Graphic Processing Unit). In the proposed SLAM system, deep neural network Mask-RCNN requires high level GPUs where the onboard computer system may not be able to provide. In \cite{Cloud-chaser}, \cite{Cloud-parallel} and \cite{Offloading-cloud}, there are mentions of different algorithms which are used to deploy a certain portion of the entire SLAM system in the cloud. For example - The Cloud Chaser \cite{Cloud-chaser} system deployed its object detection algorithm in cloud to get better performance and avoid latency issues. 
\par 
DyOb-SLAM consists of a dynamic object detector which is computationally heavier in comparison with the other modules in the system. Besides, the output of object detector is also pre-processed which results in a slower computation time, leading to weaker performance for the SLAM system. To obtain a much better performing system, we deploy the proposed SLAM system in a cloud environment. The output results are then compared to the ones resulting in the local computer.

\section{Experimental Data} \label{sec:exp}
The proposed system has been evaluated on the basis of camera pose, object motion, object speed and moving object tracking performance. At first the experiment is done in a local computer system. The experiment is done in an Intel Core i7-8700 CPU @ 3.20GHz × 12 system with a 4 GB GPU of NVIDIA Quadro P620 processor. Then the SLAM system is run in the OSCER environment. The SLAM system, at present, is only applicable for outdoor scenes, mostly with scenes having objects like - car or any other vehicles. The main dataset that has been effective to evaluate the performance is the KITTI Tracking dataset [25]. 

\subsection{KITTI Tracking Dataset}
The KITTI Tracking Dataset \cite{Vision-meets-robotics} \cite{kitti} is a dataset collection for the use of autonomous driving and mobile robotics research. The calibrated, synchronized and time-stamped dataset collection consists of real world traffic situations. The data are developed for stereo, optical flow, visual odometry/SLAM and 3D Object detection experiments. 
\par
\begin{figure}[]
    \centering
    \includegraphics[scale=0.2]{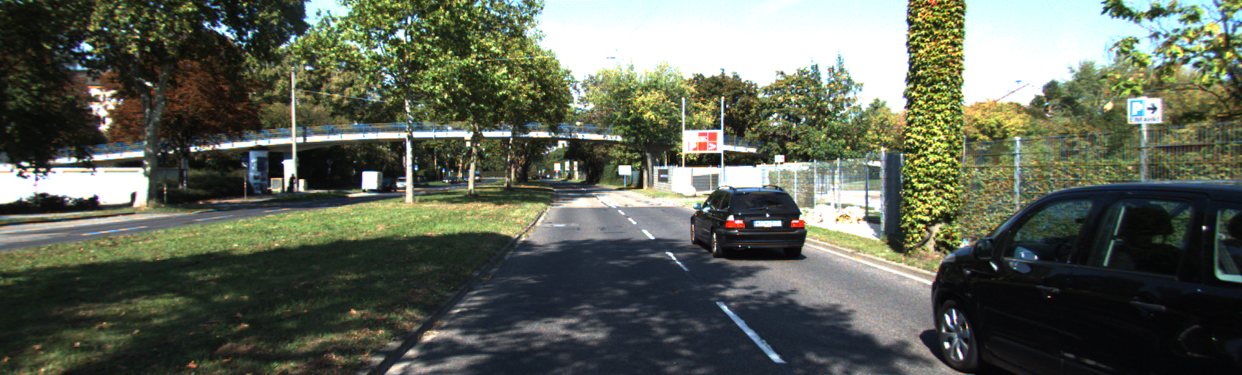}
    \caption{RGB image of a frame of \emph{kitti-0000-0013}}
    \label{figure 05}
\end{figure}

The KITTI data sequence that was chosen for this experiment is \emph{kitti-0000-0013}. This dataset contains 55 sequences with RGB images (see figure \ref{figure 05}), depth images, timestamps and ground truth information of camera and object pose. The sequences are of a simple scene where two cars driving in a road. 

The setting file for the KITTI dataset is calibrated to tune the parameters for running the SLAM system. The camera  parameters like the camera calibration and distortion parameters, the camera frames per second, frame height and width are chosen according to the dataset which have not been altered. Since the dataset is comprised of RGB-D images, the depth parameters need to be set as well. The depth value has been set for the two different features - static and dynamic. The variable \emph{ThDepth} denote the depth value for the static features and \emph{ThDepthObj} denote the depth value for dynamic features. The depth values help to differentiate the close and far points in the images. The depth map factor is also set which is a scale factor that multiplies the input depthmap.  

\subsection{Camera and Object Pose}
\begin{figure}[]
    \centering
    \includegraphics[scale=0.3]{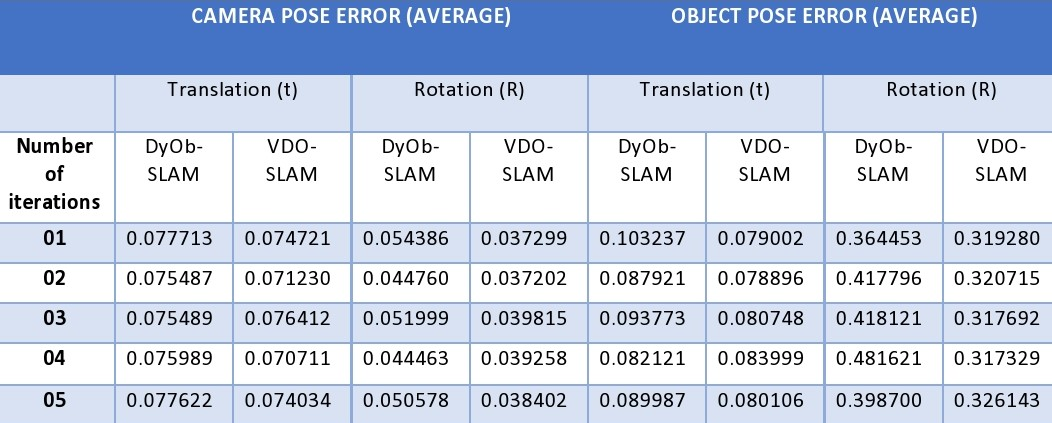}
    \caption{Table of Comparison of errors for DyOb-SLAM and VDO-SLAM}
    \label{figure 06}
\end{figure}

The main comparison of performance of the DyOb-SLAM system is established with VDO-SLAM. The table in figure \ref{figure 06} shows the average of both Camera pose error and object pose error containing both translational and rotational errors. Each sequence has been run 5 times for both DyOb-SLAM and VDO-SLAM to locate the change in the non-deterministic output. 

Let the ground truth motion transform be denoted as $\mathcal{T}$, where $\mathcal{T} \in SE(3)$ and the estimated motion transform be denoted as $\mathcal{T}_{est}$. The pose error P will generally be calculated as -
\begin{equation} \label{eq:5}
    P = \mathcal{T}_{est}^{-1}\mathcal{T}
\end{equation}
At the different camera frames and time steps, the root mean squared errors (RMSE) for both camera pose and object motion pose are computed. After all the frames are processed, the end result is an average of all the RMS errors calculated in each frames. The average pose error is calculated for both camera and object motion.  

\subsection{Object Speed}
For both DyOb-SLAM and VDO-SLAM systems, the object speed error has been evaluated. The linear velocity of each points in the object pixels are estimated. If the pose change is expressed as $H$ and $m_k$ be a point in the object pixel at frame $k$ where $m=[m_x, m_y, m_z, 1]$, then the estimated velocity can be expressed as

\begin{align} \label{eq:6}
    \mathcal{V}_{est} &= m_k - m_{k-1} \nonumber \\
     &= H_k m_{k-1} - m_{k-1} = (H_k  - I_4) m_{k-1}
\end{align}

Here, $I_4$ is an identity matrix. The equation 5.2 states that the difference of the point coordinates from frame k-1 to k at the time step gives the estimated velocity. 

\begin{figure}[]
    \centering
    \includegraphics[scale=0.25]{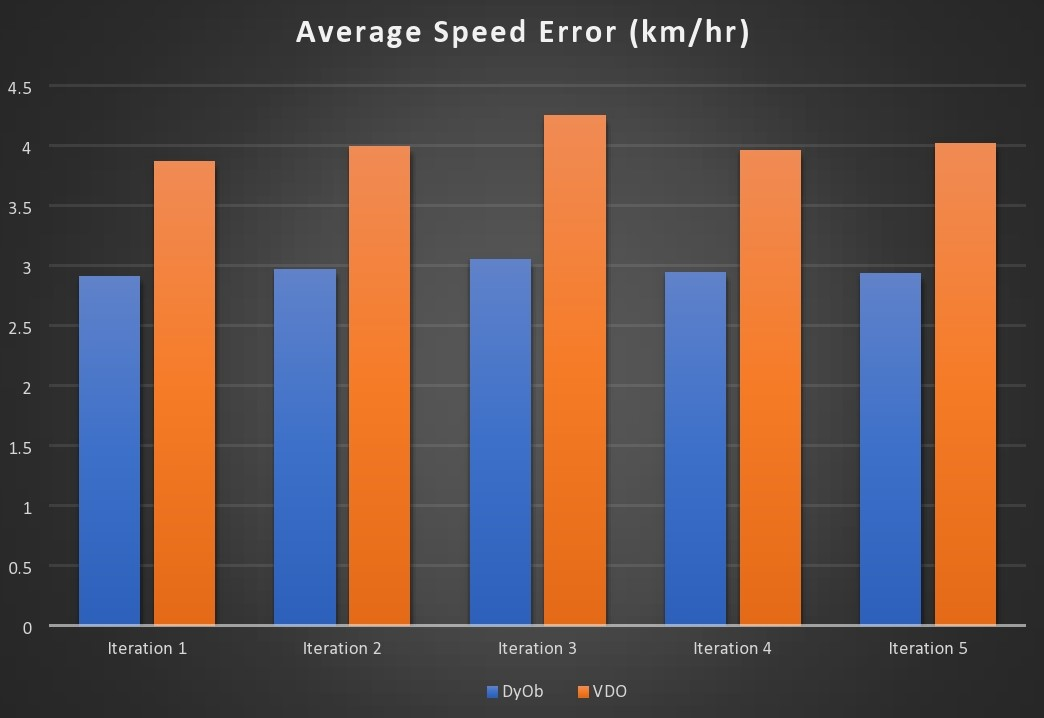}
    \caption{Average object speed error for DyOb-SLAM and VDO-SLAM}
    \label{figure 07}
\end{figure}

The figure \ref{figure 07} shows a bar chart of the average object speed error for DyOb-SLAM and VDO-SLAM at the 5 iterations. If $\mathcal{V}$ is the ground truth speed information, the velocity error - 
\begin{equation} \label{eq:8}
    V_{err} = \lvert \mathcal{V}_{est} \rvert - \lvert \mathcal{V} \rvert
\end{equation}

\begin{figure}[]
    \centering
    \includegraphics[scale=0.2]{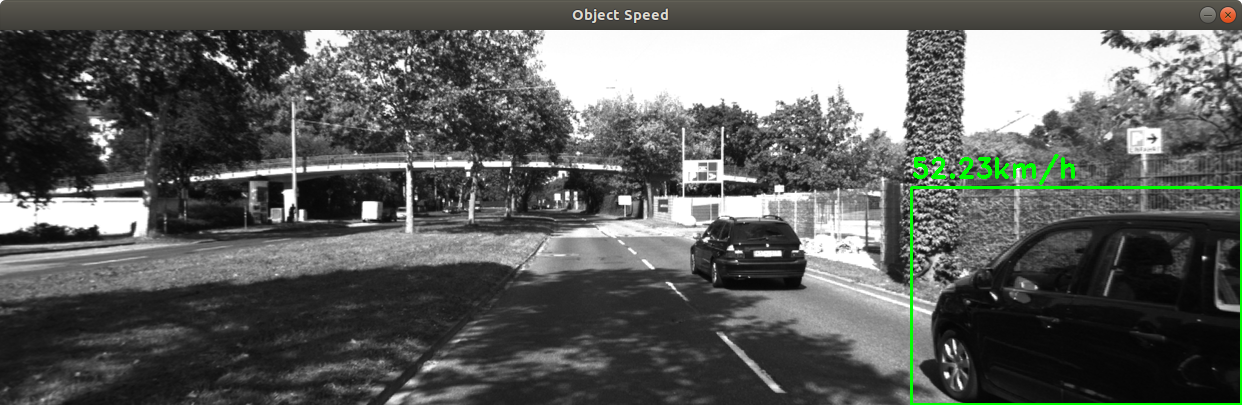}
    \caption{Object Speed Calculation in each frame}
    \label{figure obj_speed}
\end{figure}

The object speed is calculated at the end of processing every frame. In figure \ref{figure obj_speed} it is shown that the detected dynamic objects are bounded by bounding boxes and the speed of the objects are printed in the processed frame. The average of these errors are obtained as the output of the SLAM system.

\subsection{Discussion of Results}
The key difference between VDO-SLAM and DyOb-SLAM is the semantic segmentation process in the system. For VDO-SLAM, the semantic segmentation is done in a pre-processed way shaping the segmentation data according to the ground truth. On the other hand DyOb-SLAM directly uses a Mask-RCNN network which segments out the dynamic objects and later on the data is then processed (see figure \ref{figure 08}). Due to this, the average pose and speed errors are different for the two SLAM systems. 
\par 
As we can see from figure \ref{figure 06}, the proposed system DyOb-SLAM has slightly higher average pose error for both camera and object in every iterations. Nevertheless, both the systems have very low average pose error. It can be observed from the table that the camera translational error of DyOb-SLAM differ about a range of 0.003-0.005 than that of VDO-SLAM. The camera rotational error difference is seen to be about a range of 0.007-0.017 which is a bigger difference compared to the translational error difference. For object pose error, the translational error difference ranges about 0.009-0.03 and the rotational error difference ranges about 0.05-0.17. It is well observed that for DyOb-SLAM system the first iteration gave higher outputs than the next 4. It is also observed, for both camera and object poses, the rotational error is bigger compared to the translational error. 
\par 
On the other hand, for the average object speed performance, it is observed that DyOb-SLAM performs better than VDO-SLAM which can be seen from figure \ref{figure 07}. The object speed error for DyOb-SLAM gives a range of 0.96-1.20 more than the object speed error found from VDO-SLAM. The difference is very high and it can be concluded that the DyOb-SLAM system gives a satisfactory object speed error data.  
\par 
The settings file that has been obtained from the KITTI dataset has been modified for the experiment. With Mask-RCNN's segmentation information, the objects could not be categorized as multiple. The scene contains different vehicles. The neural network segments out even the static prior outlier points. To solve the matter and to only segment out the required two cars that need to be tracked, the depth value has been compromised to a limit. As a result of this, the two focused cars in the frames can be identified as separate objects, instead of one. For this reason, the pose errors in the proposed system are a bit higher than that of VDO-SLAM. Besides, if the computational energy of the two system is compared, we understand that the proposed system is more expensive than VDO-SLAM because we have used the Mask-RCNN model which functions at the same time as the Tracking and Mapping modules in the system. 

\par
The speed of the detected objects is calculated when the objects are detected and segmented out. After the dynamic objects are labelled with an object ID, the bounding boxes along with the speed calculation appear as seen in figure \ref{figure obj_speed}. Due to focusing the two vehicles at the close point portion in the frames which have been discussed before, the object bounding boxes appear as per the segmented objects. This means that when the detected objects are in close point depth scale, the bounding boxes along with the speed calculation appear accordingly. So, after the first car goes out of the depth scale, the speed of the car is not measured.

\section{CONCLUSION} \label{sec:Conclusion}
In this work, we propose the DyOb-SLAM algorithm, which is one of the SLAM systems that includes dynamic objects in its mapping and track the objects in every frames. With the help of advanced technologies like convolutional neural networks and dense optical flow algorithms, dynamic contents can be accurately detected and used for further processing. The
back-end of the DyOb-SLAM system can provide more optimized data for which
an optimized sparse map of static keypoints in the environment can be obtained, along with
a global map of camera and object trajectories with subsequent processing of each frame.
To obtain an even better SLAM processing system, a cloud environment can be used to process data in the main processing unit and the cloud at the same time. In that way, the pose estimation error can be lowered and a faster driven output can be obtained.

\addtolength{\textheight}{-12cm}   




\section*{ACKNOWLEDGMENT}

We would like to thank Mr. Dharmendra, Head Of Department United College of Engineering and Research, for the constant encouragement towards the realization of this work.


\bibliographystyle{ieeetr}
\bibliography{ref}

\end{document}